\documentclass[lettersize,journal]{IEEEtran}
\usepackage{amsmath,amsfonts}
\usepackage{algorithmic}
\usepackage{algorithm}
\usepackage{array}
\usepackage[caption=false,font=normalsize,labelfont=sf,textfont=sf]{subfig}
\usepackage{textcomp}
\usepackage{stfloats}
\usepackage{url}
\usepackage{verbatim}
\usepackage{graphicx}
\usepackage{cite}
\usepackage{tikz}
\usepackage{booktabs}
\usepackage{multirow}
\usepackage{colortbl}

\newcommand*{\rom}[1]{\expandafter\@slowromancap\romannumeral #1@}
\newcommand*{\circled}[2][]{\tikz[baseline=(C.base)]{
    \node[inner sep=0pt] (C) {\vphantom{1g}#2};
    \node[draw, circle, inner sep=1pt, yshift=1pt] 
        at (C.center) {\vphantom{1g}};}}

\begin{document}

\title{Diversity-boosted Generalization-Specialization Balancing for Zero-shot Learning}

\author{Yun~Li,
        Zhe~Liu,
        Xiaojun~Chang~\IEEEmembership{Senior~Member,~IEEE},
        Julian McAuley,
        and~Lina~Yao~\IEEEmembership{Senior~Member,~IEEE}%
\thanks{Y. Li is with the School of Computer Science and Engineering, University of New South Wales, Sydney, NSW 2052, Australia (e-mail: yun.li5@unsw.edu.au). }
\thanks{Z. Liu is with (1) the Jiangsu Provincial Engineering Laboratory of Pattern Recognition and Computational Intelligence, Jiangnan University, Wuxi, China 214122, and also with (2) the School of Computer Science and Engineering, University of New South Wales, Sydney, NSW 2052. (e-mail: zheliu912@gmail.com)}
\thanks{X. Chang is with the Faculty of Engineering and Information Technology, University of Technology Sydney, Sydney, NSW 2007, Australia (e-mail: xiaojun.chang@uts.edu.au).}%
\thanks{J. McAuley is with the School of Computer Science and Engineering, University of California San Diego, San Diego, California, United States (e-mail: jmcauley@eng.ucsd.edu).}
\thanks{L. Yao is with CSIRO's Data61 and University of New South Wales, Sydney, NSW 2052, Australia (e-mail: lina.yao@data61.csiro.au).}}

\markboth{}%
{Li \MakeLowercase{\textit{et al.}}: Diversity-boosted Generalization-Specialization Balancing for Zero-shot Learning}


\maketitle

\begin{abstract}
Zero-Shot Learning (ZSL) aims to transfer classification capability from seen to unseen classes. Recent methods have proved that generalization and specialization are two essential abilities to achieve good performance in ZSL. However, focusing on only one of the abilities may result in models that are either too general with degraded classification ability or too specialized to generalize to unseen classes. In this paper, we propose an end-to-end network, termed as BGSNet, which equips and balances generalization and specialization abilities at the instance and dataset level. Specifically, BGSNet consists of two branches: the Generalization Network (GNet), which applies episodic meta-learning to learn generalized knowledge, and the Balanced Specialization Network (BSNet), which adopts multiple attentive extractors to extract discriminative features and achieve instance-level balance. A novel self-adjusted diversity loss is designed to optimize BSNet with redundancy reduced and diversity boosted. We further propose a differentiable dataset-level balance and update the weights in a linear annealing schedule to simulate network pruning and thus obtain the optimal structure for BSNet with dataset-level balance achieved. Experiments on four benchmark datasets demonstrate our model's effectiveness. Sufficient component ablations prove the necessity of integrating and balancing generalization and specialization abilities.
\end{abstract}


\section{Introduction}
Humans can easily accumulate past knowledge to perceive novel concepts. Inspired by this, Zero-Shot Learning (ZSL) is proposed to perform inference over novel classes whose samples are unseen during training. The bridge between seen and unseen classes is the shared semantic attributes that describe the visual appearance, e.g., a shared attribute \textit{stripes} can link \textit{tiger} (seen) and \textit{zebra} (unseen). A more rigorous extension of ZSL is Generalized Zero-Shot Learning (GZSL), \textcolor{black}{which further requires retaining the ability to classify seen classes~\cite{xian2019zero}}.
Zero-shot learning and generalized zero-shot learning have attracted plenty of interest recently~\cite{wang2018zero,verma2017simple,zhu2019generalized,gao2020zero,yang2020simple,verma2020meta} for their potential to reduce expensive annotation costs. 


A typical scheme of ZSL/GZSL is to learn the visual representations from images, project visual and semantic embeddings to a common space, \textcolor{black}{and then perform a nearest neighbor search in the space for classification}~~\cite{zhu2019semantic,xu2020attribute,sylvain2019locality,li2021attribute,liu2018generalized,wang2019tafe,liu2020attribute,ye2019progressive,hu2020semantic}. Reviewing these successful methods, we find that 
the abilities of generalization, i.e., how well the learned representation can be transferred to unseen classes, and specialization, i.e., extracting discriminative features, are two principles that lead to good performance. 

Targeting better generalization, recent works explore meta-learning to learn more generalized knowledge~\cite{li2021attribute,liu2021task,liu2021isometric,verma2020meta,meta_zeroshot1,meta_zeroshot2,meta_zeroshot3,meta_zeroshot4}, or implicitly learn semantic-visual correlations on unseen classes by synthesizing unseen samples~\cite{zhang2019triple,gao2020zero,zhang2021modality,chen2018zero,zhu2019generalized,felix2018multi,yang2020simple}. With generalization ability, models can transfer knowledge to unseen classes. But to classify images, specialization is indispensable. Some efforts adopt attention mechanisms to extract discriminative features~\cite{wang2015multiple,zhang2016picking,xie2019attentive,xie2020region,huynh2020fine}, or directly localize distinct regions by semantic guidance~\cite{zhu2019semantic,xu2020attribute} or reinforcement learning~\cite{li2022entropy,ge2021semantic,liu2021rethink}.
However, most of them fail to consider the two abilities simultaneously, causing their models to be either too general to classify highly-similar images or too specialized and thus overfit to seen classes.

This phenomenon exists not only at the instance level, i.e., when classifying a specific image, but also at the dataset level. For example, 
AwA2~\cite{xian2019zero} and SUN~\cite{patterson2012sun} are two widely-used ZSL benchmark datasets, containing 32 coarse-grained and 717 fine-grained classes, respectively. Intuitively, AwA2 has a higher demand for generalized knowledge, while SUN requires a model that strengthens discriminative features. How to design a model that can fit datasets based on their characteristics remains to be solved.

\begin{figure*}[htb]
    \centering
    \includegraphics[width=0.85\textwidth]{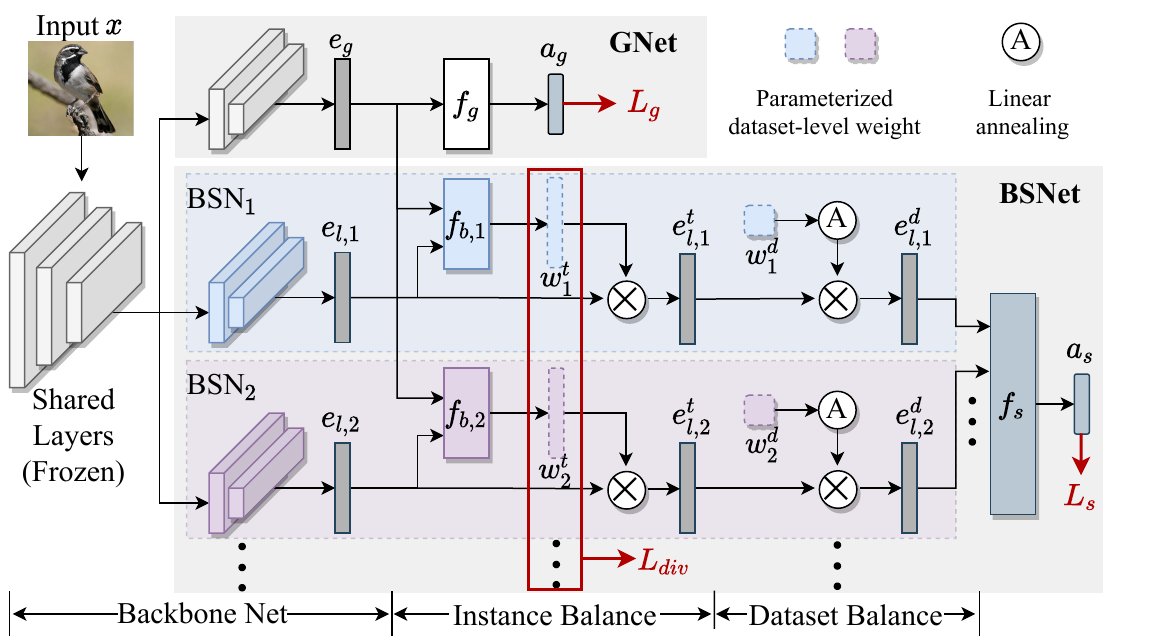}
\caption{
Overview of BGSNet. 
BGSNet consists of two branches: GNet and BSNet. 
For GNet, the input image $x$ is fed into the Backbone Net to extract the visual embedding $e_{g}$. Then, the predictor $f_{g}$ embeds $e_{g}$ to the semantic space as $a_g$. 
For BSNet, $x$ is passed through $N$ sub-modules \{BSN$_i\}_1^N$, where visual extraction, instance-level balancing, and dataset-level balancing are performed. BSN$_{i}$ shares most backbone layers with GNet while adopting different tail layers to extract specialized visual embedding $e_{l,i}$. 
$e_{l,i}$ and $e_g$ are further used to generate instance-level weights $w^{t}_{i}$ to conduct instance-level balancing. 
Then, dataset-level balancing is performed by using parameterized weight $w^{d}_{i}$ in a linear annealing schedule. 
All the balanced embeddings $e^{d}_{l,i}$ are combined to get the final semantic embedding $a_{s}$.
Classification loss $L_{g}$ and $L_{s}$, and diversity loss $L_{div}$ are leveraged for training.
Note that shared layers in Backbone Net are frozen during the optimization.
}
\label{model}
\end{figure*}

In this work, we propose a two-branch network, dubbed BGSNet, \textcolor{black}{to equip generalization and specialization abilities at the same time and to balance the two abilities at the instance- and dataset-level}. The two branches are a Generalization Network (GNet), which adopts episodic meta-learning to learn transferable knowledge, and a Balanced Specialization Network (BSNet), composed of multiple attentive feature extractors focusing on discriminative visual information. 
Unlike existing methods, we generate attention weights based on both generalized and specialized visual embeddings. Thus attention can enhance distinctive features and also balance the two abilities by adjusting weights of BSNet.
We design a dynamic diversity loss to optimize instance-level balance. 
This novel loss is self-adjusted based on weight distributions of visual channels and BSNet sub-modules. It has a self-calculated margin to adjust the optimization purposes during training, and ultimately reduce redundancy and increase specialization diversity.
Additionally, we apply differentiable dataset-level weights to further balance the two abilities. The dataset-level weights are updated in a linear annealing schedule and will finally be binarized to simulate network pruning. Thus it can find the approximate-best network structure. 

\noindent \textbf{Contributions.} In summary, we make the following contributions in this paper:
\begin{itemize}
    \item We present BGSNet for zero-shot learning and generalized zero-shot learning. BGSNet combines and balances generalization and specialization abilities, thus retaining the discrimination power and providing enough flexibility to learn novel concepts simultaneously.
    \item We design a dynamic diversity loss for the specialization branch to avoid channel and modular redundancy and boost diversity. We also prove its correctness theoretically and  effectiveness experimentally.
    \item We conduct extensive experiments on four benchmark datasets in both ZSL and GZSL settings, \textcolor{black}{and BGSNet consistently outperforms or perform comparably to the SOTA}, which demonstrates the effectiveness of BGSNet. We further conduct ablation studies of each component in our model, and analyse the learned instance- and dataset-level balance.
\end{itemize}

\section{Method}

\noindent \textbf{Task definitions and notations.} 
Suppose that training data $\mathcal{S}=\{(x, y, a)|x\in{X^{S}},y\in{Y^{S}},a\in{A^{S}}\}$ from seen classes (classes with labeled samples) are given, where $x\in{X^{S}}$ is an image with its label $y\in{Y^{S}}$, and $a\in{A^{S}}$ represents the corresponding attribute of class $y$ (or other semantic side information, such as word embeddings of text descriptions~\cite{akata2015evaluation,pennington2014glove}). For ZSL, given test set $\mathcal{U}=\{(x, y, a)|x\in{X^{U}},y\in{Y^{U}},a\in{A^{U}}\}$ from unseen classes, we aim to predict the label $y\in{Y^{U}}$ for each image $x\in{X^{U}}$. The class sets of training and testing are disjoint, i.e.,~$Y^{S}\cap{Y^{U}}= \emptyset$. Meanwhile, for GZSL, the testing class sets are expanded to ~$\mathcal{Y} = Y^{S}\cup{Y^{U}}$, i.e., testing images can be from both seen and unseen classes.

\subsection{Overview}

As Figure~\ref{model} shows, BGSNet consists of two branches: the Generalization Net (GNet) and the Balanced Specialization Net (BSNet). 
GNet (§~\ref{sec:gnet}) adopts meta-learning to learn more generalized knowledge that is commonly shared across classes and can be easily transferred to unseen classes. 
BSNet (§~\ref{sec:snet}) maximizes the specialization capability to discover discriminative features for classification. 
It also balances the generalization and specialization abilities of BGSNet by adjusting the weights of BSNet at the instance- and dataset- level. 
The dataset-level weights also play a role in automatically finding the best network structure of BSNet in a linear annealing schedule. 
Both branches are trained with classification loss, i.e., $L_{g}$ and $L_{s}$, respectively. 
An additional diversity loss $L_{\mathit{div}}$ is proposed to optimize the instance-level weights across sub-modules BSN$_{i}$ in BSNet to enrich the diversity.

\subsection{Generalization Branch: GNet}\label{sec:gnet}

GNet aims to extract features that easily generalize to unseen classes rather than reflect subtle differences in seen classes. Given an input image $x$, Backbone Net (i.e., ResNet101~\cite{he2016deep} in our experiments) embeds $x$ to a visual embedding $e_{g}$, and the predictor $f_{g}$ further projects $e_{g}$ into the semantic space as $a_g$. GNet is optimized by the attribute-incorporated CrossEntropy loss $\mathcal{L}_{g}$ to improve the compatibility between $a_g$ and its true attribute:
\begin{equation}
\resizebox{.91\linewidth}{!}{$
    \displaystyle
    \mathcal{L}_{g}  = \mathit{CrossEntropy}(a_{g},y)
     = -\log\frac{\exp(a_{g}^{T}\phi(y))}{\sum_{\hat{y}\in Y^{S}}\exp(a_{g}^{T}\phi(\hat{y}))}
$}
\end{equation}
where $y$ denotes the label of $x$; $\phi(y)$ is the attribute of $y$. 

\textcolor{black}{In contrast to standard ZSL methods with direct-training strategy, we further improve the generalization capability of GNet, restrain overfitting towards seen classes, and alleviate domain shift~\cite{verma2020meta,liu2020attribute,liu2021isometric} following the idea of episodic meta-learning~\cite{finn2017model}}. In detail, we first train GNet for adequate epochs to provide a preliminary classification ability and then freeze the Backbone Net and train $f_g$ in an episode-wise manner. 

In each episode, we sample batches of tasks. 
Each task contains a training set $D^{j}_{\mathit{tr}}$ and a validation set $D^{j}_{val}$. Their samples belong to two disjoint class sets $Y^{j}_{tr}$ and $Y^{j}_{val}$, respectively ($Y^{j}_{tr}, Y^{j}_{val} \subseteq Y^{S}$). We first use $D^{j}_{tr}$ to virtually optimize $f_{g}$ to get a virtual parameter $\theta_{g}^{'}(D^{j}_{tr})$, and then we minimize the cumulative validation loss of $D^{j}_{val}$ calculated based on virtual $f_{g}$ to truly update $f_{g}$. We summarize the meta optimization for each episode as follows:
\begin{gather}
\theta_{g}^{'}(D^{j}_{tr}) \leftarrow \theta_{g}-\alpha\bigtriangledown_{\theta_{g}}\mathcal{L}_{\mathcal{D}^{j}_{tr}}^{g}(\theta_{g})
\\
\theta_{g}^{*}\leftarrow\theta_{g}-\beta\sum_{j}\bigtriangledown_{\theta_{g}}\mathcal{L}^{g}_{\mathcal{D}^{j}_{val}}(\theta^{'}_{g}(D^{j}_{tr}))
\end{gather}
where $\theta_{g}^{'}(D^{j}_{tr})$ are the virtual parameters of $f_{g}$; $\theta_{g}^{*}$ are the updated parameters; $\alpha$ and $\beta$ are learning rates.

\subsection{Balanced Specialization Branch: BSNet}\label{sec:snet}

The features learned by GNet may be too generalized and thus may not be sufficient to classify similar images, especially those from fine-grained datasets. To improve the specialization ability of our model, we propose the BSNet focusing on capturing discriminative features. To avoid harming generalization ability when strengthening specialization and to achieve dataset-specific and instance-specific balance, we further equip the BSNet with the instance- and dataset-level balance.

As shown in Figure~\ref{model}, BSNet is composed of $N$ sub-modules BSN$_i$, where $i\in [1, N]$. BSN$_i$ first projects the input image into visual embeddings $e_{l,i}$. To save computation cost, we split the Backbone Net into two parts: shared layers and tail layers (we use the last nine convolutional layers and the pooling layer as tails in our experiments). The shared layers are frozen during training. Only tail layers are optimized. 
Instance- and dataset-level balance are then performed. 

\subsubsection{Instance-level Balance}\label{i-balance}

For each sub-module, the instance-level balance generator $f_{b,i}$ takes $e_{l,i}$ and $e_g$ as inputs, evaluates the generalization and specialization preference when classifying a specific image, and yields instance-level weights $w^{t}_{i}$. Specifically, the instance-weighted embedding $e^{t}_{l,i}$ is obtained by:

\begin{equation}
e^{t}_{l,i}=w^{t}_{i} \otimes e_{l,i}=f_{b,i}(e_{l,i},e_{g}) \otimes e_{l,i}
\end{equation}
where $e_{l,i}\in R^{1\times K}$, $w^{t}_{i}\in (0,1)^{1\times K}$, and $K$ is the channel number; $\otimes $ indicates a channel-wise product.

$w^{t}_{i}$ have two functions: 
1) serving as an attention layer, allowing BSN$_{i}$ to focus on important parts and then improve the specialization ability; 
2) balancing the generalization (GNet) and specialization (BSNet) abilities by adjusting the weights of BSN$_i$.

Realizing the functions requires: (\uppercase\expandafter{\romannumeral1}) BSNet learns novel and discriminative knowledge that GNet cannot capture. 
Then, for a sub-module $BSN_i$, $w^{t}_{i}$ should follow a distribution such that: most of the weights are close to 0, and weights of special information are close to 1.
In this way, we can reduce redundancy, strengthen unique visual information, and balance $e_g$ and $e_{l,i}$ by adjusting $e_{l,i}$ with $w^{t}_{i}$. (\uppercase\expandafter{\romannumeral2}) For the first function, we further expect that $w^{t}_{i}$ of different sub-modules $\{BSN_{i}\}_{1}^{N}$ focus on different aspects, i.e., for the $k^{th}$ channel ($k\in[1,K]$), the weights are diverse across sub-modules. Then Requirement \uppercase\expandafter{\romannumeral1} can be extended to: for $k^{th}$ channel, most sub-modules have $w^t_{i,k}$ close to 0, and only sub-modules attending this channel have weights near 1.
Intuitively, the divergent distribution across different $w^{t}_{i}$ enriches the diversity of specialization and thus facilitates classification.

Therefore, we propose a diversity loss $L_{div}$ to realize the functions and meet the requirements as follows:
\begin{equation}\label{divergence_loss}
    \mathcal{L}_{div}=\sum_{k\in[1,K]}\sum_{i\in[1,N]}[w^{t}_{i,k}\widehat{w}^{t}_{i,k}-mrg_{k}(w^{t}_{i,k}+\widehat{w}^{t}_{i,k})]
\end{equation}
where $\widehat{w}^{t}_{i,k}=\max_{i\neq n} w^{t}_{n,k}$ denotes the maximum value of other weights at $k^{th}$ channel and $mrg_{k}$ is a self-calculated margin for $k^{th}$ channel.

$L_{dis}$ first accumulates the channel-wise loss and then the module-wise loss. Therefore, for each channel, we can sort $w^{t}_{i,k}$ to have the ordered weights without influencing the loss calculation: $w^{t}_{k}=\{w^{t}_{1,k}\leq w^{t}_{2,k} \leq ... \leq w^{t}_{N,k}\}$. When calculating divergence loss for $w^{t}_{i,k}$, it will propagate gradients to $w^{t}_{i,k}$ and $\widehat{w}^{t}_{i,k}$. Therefore, each calculation of the instance-level divergence loss will produce a pair of gradients as follows:
\begin{gather}
\bigtriangledown \mathcal{L}_{div}(w^{t}_{i,k})_{w^{t}_{i,k}}=\widehat{w}^{t}_{i,k}-mrg_{k}\\
\bigtriangledown \mathcal{L}_{div}(w^{t}_{i,k})_{\widehat{w}^{t}_{i,k}}=w^{t}_{i,k}-mrg_{k}
\end{gather}
where $\bigtriangledown \mathcal{L}_{div}(w^{t}_{i,k})_{w^{t}_{i,k}}$ and $\bigtriangledown \mathcal{L}_{div}(w^{t}_{i,k})_{\widehat{w}^{t}_{i,k}}$ denote the gradients propagated to optimize $w^{t}_{i,k}$ and $\widehat{w}^{t}_{i,k}$, respectively.

Then we can summarize gradients of $\mathcal{L}_{div}(w^{t}_{k})$ as follows:
\begin{gather}
    \bigtriangledown \mathcal{L}_{div}(w^{t}_{k})_{w^{t}_{i,k}}=w^{t}_{N,k}-mrg_{k} \quad (i\in[1,N-2])\\ 
        \bigtriangledown \mathcal{L}_{div}(w^{t}_{k})_{w^{t}_{N-1,k}}=2*(w^{t}_{N,k}-mrg_{k}) \label{equ9} \\
        \bigtriangledown \mathcal{L}_{div}(w^{t}_{k})_{w^{t}_{N,k}}=\sum_{1}^{N-1} w^{t}_{i,k}+ w^{t}_{N-1,k}-N*mrg_{k} \label{equ10}
\end{gather}
where $N\geq 3$, otherwise only Equations~\ref{equ9}-\ref{equ10}  are established.

From the above equations, we can easily find that:
\begin{gather}
    \bigtriangledown \mathcal{L}_{div}(w^{t}_{k})_{w^{t}_{i,k}}\propto w^{t}_{N,k}-mrg_{k} \quad (i\in[1,N-1])\\
    \bigtriangledown \mathcal{L}_{div}(w^{t}_{k})_{w^{t}_{N,k}}\leq N*(\Bar{w}^{t}_{k}-mrg_{k}) \propto \Bar{w}^{t}_{k}-mrg_{k}
\end{gather}
where $\Bar{w}^{t}_{k}$ is the mean value of weights $w^{t}_{k}$ in the $k^{th}$ channel.

To meet the two requirements, we want to optimize most weights (i.e., $\{w^{t}_{i,k}\}^{N-1}_1$) towards 0, and $w^{t}_{N,k}$ towards 1. Therefore, we need $\bigtriangledown L_{div}(w^{t}_{k})_{w^{t}_{i,k}}>0$ and $\bigtriangledown L_{div}(w^{t}_{k})_{w^{t}_{N,k}}<0$. Then, the upper and lower bounds of $mrg_{k}$ to realize our goal are as follows:
\begin{equation}
    mrg_{k}\in(\Bar{w}^{t}_{k},w^{t}_{N,k})
\end{equation}
where $\Bar{w}^{t}_{k}$, $w^{t}_{N,k}$ are the mean value and the largest weight in $k^{th}$ channel. Note that $mrg_{k}$ exists only if $w^{t}_{1,k}\neq w^{t}_{N,k}$ in the ordered $k$ channel weights; otherwise $mrg_{k}\in \emptyset$. Therefore, we initialize the weights with different values.

Based on the bounds, we design self-calculated $mrg_{k}$ as:
\begin{equation}
    mrg_{k} = max(\bar{w}^{t}_{k}+\epsilon,(1-\frac{1}{1+\mathit{eph}})w^{t}_{N,k})
\end{equation}
where $\epsilon$ is a small value (e.g., $1e^{-6}$) to meet the lower bound (we add $\epsilon$ only when $\bar{w}^{t}_{k}+\epsilon < w^{t}_{N,k}$); $\mathit{eph}$ is the epoch index. 

During training, at the early stage when $\mathit{eph}$ is small, $mrg_{k}$ tends to be near the lower boundary. $\mathcal{L}_{div}$ will decrease most weights to 0. As $\mathit{eph}$ increases, $mrg_{k}$ approaches the upper boundary. $\mathcal{L}_{div}$ increases weights of important channels to 1. With this channel-wise self-calculated margin, $\mathcal{L}_{div}$ in Eq.~\ref{divergence_loss} can reduce redundancy and boost diversity. 

\subsubsection{Dataset-level Balance}\label{d-balance}

This part aims to balance GNet and BSNet at dataset level and searches the best structure of BSNet according to dataset characteristics, e.g, fine-grained or coarse-grained. Given $e^{t}_{l,i}$, the balanced embedding $e^{d}_{l,i}$ is obtained by (adapted from HAT~\cite{serra2018overcoming}): 

\begin{equation}
   e^{d}_{l,i}= e^{t}_{l,i} \otimes \Bar{w}^{d}_{i}= e^{t}_{l,i} \otimes \mathit{Sigmoid}(s*w^{d}_{i})
\end{equation}
where $w^{d}_{i}$ is a parameterized dataset-level weight and is differentiable. 
$\mathit{Sigmoid}$ activates $w^{d}_{i}$ to fall into (0,1), and $s$ is used as a scaling factor to adjust its sharpness. During training, $s$ is calculated in a linear annealing schedule:

\begin{equation}
  \resizebox{.85\linewidth}{!}{$
    \displaystyle
  s = \circled{A}(\mathit{eph}) = \frac{1}{s_{max}}+(s_{max}-\frac{1}{s_{max}})*\frac{eph-1}{eph_{max}-1}
  $}
\end{equation}
where $s_{max} \gg 1$ is a hyper-parameter controlling the annealing schedule; $\mathit{eph_{max}}$ is a pre-defined maximum epoch. 

At first, since $s_{max} \gg 1$, $s$ is close to 0, allowing $\mathit{Sigmoid}$ to activate $w^{d}_{i}$ in a uniform way that $s$ approximately equals to 0.5. As $\mathit{eph}$ increases, $s$ increases, and $\Bar{w}^{d}_{i}$ will be gradually binarized. 
This design has two benefits: 1) it avoids premature deactivation of BSN$_i$ to allow fully training; 2) the final binarized weight $\Bar{w}^{d}_{i}$ can be regarded as an approximate pruning of BSNet, i.e., it finds the approximately best structure for BSNet. 

\begin{table}[htb]
\centering
\caption{Overall comparisons in ZSL. The performance is evaluated by average per-class Top-1 accuracy (\%). * indicates end-to-end methods or methods using fine-tuned ResNet-101 as backbones. Best results are in bold.}
\begin{tabular}{l|c|c|c|c}
\toprule
Method &~~SUN~~ & ~~CUB~~ & ~~aPY~~ & ~~AwA2~~ \\
\midrule
TCN~\cite{jiang2019transferable}&61.5&59.5&38.9&\textbf{71.2}\\
PREN~\cite{ye2019progressive} & 60.1 & 61.4 & - & 66.6 \\
*LFGAA~\cite{Liu_2019_ICCV}&61.5&67.6&-&68.1\\
*SGMA~\cite{zhu2019semantic}&- &71.0&-&68.8\\
*AREN~\cite{xie2019attentive}&60.6&71.5&39.2&67.9\\
TVN~\cite{zhang2019triple} & 59.3 & 54.9 & 40.9 & 68.8\\
Zero-VAE-GAN~\cite{gao2020zero} & 58.5 & 51.1 & 34.9 & 66.2\\
*APN~\cite{xu2020attribute}&60.9&71.5&-&68.4\\
*DAZLE~\cite{huynh2020fine} &59.4&66.0&-&67.9\\
*MIANet~\cite{zhang2021modality}&60.5&57.9&41.2&69.0\\
AMAZ~\cite{li2021attribute}&60.7&68.9&-&68.2\\
SCILM~\cite{ji2021semantic}&62.4&52.3&38.4&\textbf{71.2}\\
\hline
\textbf{Ours} *BGSNet  & \textbf{63.9} & \textbf{73.3} &  \textbf{43.8} & 69.1 \\
\textcolor{black}{\textbf{Ours} *Baseline}  & \textcolor{black}{60.2} &  \textcolor{black}{68.7}&  \textcolor{black}{38.5} & \textcolor{black}{65.5} \\
\bottomrule
\end{tabular} 
\label{table zsl}
\end{table}

\textbf{Training objectives of BSNet.} After passing through visual extraction, instance- and dataset- level balance in BSN$_i$, weighted embeddings $\{e^d_{l,i}\}_1^N$ are concatenated as $e_s \in R^{N\times K}$ and fed into the overall predictor $f_{s}$ to get the specialized semantic embedding $a_{s}$: $a_{s} = f_{s}(e_s)$. We adopt the classification loss $\mathcal{L}_{s}$ to improve the semantic compatibility: 

\begin{equation}
\resizebox{.85\linewidth}{!}{$
    \displaystyle
    \mathcal{L}_{s}  = \mathit{CrossEntropy}(a_{s},y)
     = -\log\frac{\exp(a_{s}^{T}\phi(y))}{\sum_{\hat{y}\in Y^{S}}\exp(a_{s}^{T}\phi(\hat{y}))}
$}
\end{equation}

We train BSNet including all sub-modules $\{BSN_{i}\}^K_1$ and  $f_{s}$ simultaneously with the following objective function:
\begin{equation}
\mathcal{L}_{BSNet} = \mathcal{L}_{s}+\eta \mathcal{L}_{div}
\end{equation}
where $\eta$ is a hyper-parameter.

\subsection{Training and Inference}\label{sec:training}
\subsubsection{Training} We first fine-tune the Backbone Net, and then freeze the Backbone and optimize GNet and BSNet separately. GNet is optimized in a meta-manner. For BSNet, we design two training strategies: \textit{Parallel}, which trains instance- and dataset-level balance simultaneously, and \textit{Sequential}, which excludes instance-level at first to train BSNet with only dataset-level balance for $\mathit{eph}_{max}$ epochs, and then includes instance-level balance to train the whole BSNet. Training in parallel can encourage instance- and dataset-level weights to cooperate better. But gradients of $\mathit{Sigmoid}$ function are unstable due to the schedule of $s$~\cite{yan2021dynamically}. Thus training sequentially can avoid that the unstable optimization of dataset-level balance impairs the optimization of instance-level balance.

\subsubsection{Inference} We use the fusion of the generalized and specialized predictions for inference: $\bar{a} = a_{g}+a_{s}$. \textcolor{black}{We set the fusion weight to 1 because the balance between the two branches is already achieved at the instance and dataset level.}
For ZSL, given an image $x$, we take the class with the highest fusion compatibility as the final prediction:
\begin{equation}
    y^{U}=\mathop{\arg\max}\bar{a}^{T}\phi(c)_{c \in Y^{U}}
\end{equation}

For GZSL, to eliminate the bias towards seen classes, we adopt Calibrated Stacking (CS)~\cite{chao2016empirical} to 
reduce
the confidence of seen classes by a pre-defined constant $\delta$. The final prediction is: 
\begin{equation}
    y^{U\cup S}=\mathop{\arg\max} <\bar{a}^{T}\phi(c)_{c \in Y^{U}}, \bar{a}^{T}\phi(c)_{c \in Y^{S}}-\delta >
\end{equation}

Note that we set $s=s_{max}$ for inference, i.e., we use the approximately pruned BSNet.  

\section{Experiments}


\begin{table*}[htb]
\centering
\caption{Overall comparisons in GZSL. The performance are evaluated by average per-class Top-1 accuracy (\%) on seen classes (S), unseen classes (U), and their harmonic mean (H). * indicates end-to-end methods or methods using fine-tuned ResNet-101 as backbones. Best results are in bold.}
\begin{tabular}{l|ccc|ccc|ccc|ccc}
\toprule
\multirow{2}{*}{Method} & \multicolumn{3}{c}{SUN} & \multicolumn{3}{c}{CUB} & \multicolumn{3}{c}{aPY} & \multicolumn{3}{c}{AwA2} \\\cmidrule{2-13}
&~~U~~ & ~~S~~ &~~H~~ & ~~U~~ & ~~S~~ & ~~H~~ & ~~U~~ & ~~S~~ & ~~H~~& ~~U~~ & ~~S~~ & ~~H~~\\
\midrule
TCN~\cite{jiang2019transferable}&31.2 &37.3 &34.0&52.6 &52.0 &52.3&24.1 &64.0 &35.1&61.2 &65.8 &63.4\\ 
PREN~\cite{ye2019progressive} & 35.4 & 27.2 & 30.8 & 35.2 & 55.8 & 43.1 & - & - & - & 32.4 & 88.6 & 47.4 \\
PQZSL~\cite{li2019compressing}&35.1 &35.3 &35.2&43.2 &51.4 &46.9&27.9 &64.1 &38.8&31.7 &70.9 &43.8 \\ 
*LFGAA~\cite{Liu_2019_ICCV}& 20.8 & 34.9 & 26.1 & 43.4 & \textbf{79.6} & 56.2 & - & - & - & 50.0 & \textbf{90.3} & 64.4\\
*SGMA~\cite{zhu2019semantic}& - & - & - & 36.7 & 71.3 & 48.5 & - & - & - & 37.6 & 87.1 & 52.5\\
*AREN~\cite{xie2019attentive}& 40.3 & 32.3 & 35.9 & 63.2 & 69.0 & 66.0 & 30.0 & 47.9 & 36.9 & 54.7 & 79.1 & 64.7\\
TVN~\cite{zhang2019triple}  & 22.2 & 38.3 & 28.1 & 26.5 & 62.3 & 37.2 & 16.1 & 66.9 & 25.9 & 27.0 & 67.9 & 38.6\\
Zero-VAE-GAN~\cite{gao2020zero} & 44.4&30.9&36.5&41.1&48.5&44.4&30.8&37.5&33.8&56.2&71.7&63.0\\
*VSG-CNN~\cite{geng2020guided} &30.3 &31.6 &30.9&52.6 &62.1 &57.0&22.9 &\textbf{66.1} &34.0&60.4 &75.1 &67.0\\
*APN~\cite{xu2020attribute}& 41.9 & 34.0 & 37.6 & \textbf{65.3} & 69.3 & \textbf{67.2} & - & - & - & 56.5 & 78.0 & 65.5\\
*DAZLE~\cite{huynh2020fine} & 24.3 &\textbf{52.3} &33.2&59.6 &56.7 &58.1&-&-&-&\textbf{75.7}& 60.3 &67.1\\
*MIANet~\cite{zhang2021modality}&22.2 &35.6 &27.4&33.3 &49.5 &39.9& 27.6 &55.8 &37.0&43.7 &70.2 &53.3\\
AMAZ~\cite{li2021attribute}&42.0 &35.1 &38.3&58.2 &55.7 &56.9&-&-&-&60.1 &69.2 &64.3 \\
SCILM~\cite{ji2021semantic}&24.8&32.6&28.2&24.5&54.9&33.8&22.8&62.7&33.4&48.9&77.8&60.1\\
\hline
\textbf{Ours} *BGSNet  &\textbf{45.2}	&34.3	&\textbf{39.0} & 60.9&	73.6&	66.7 & \textbf{31.0}&	\textcolor{black}{54.9}&	\textbf{39.7} & 61.0&	81.8&	\textbf{69.9}\\
\bottomrule
\end{tabular} 
\label{table gzsl}
\end{table*}

\subsection{Experiment settings and implementation details}

We evaluate our model on four widely-used benchmark datasets: SUN~\cite{patterson2012sun}, CUB~\cite{welinder2010caltech}, aPY~\cite{farhadi2009describing}, and AwA2~\cite{xian2019zero}.
SUN and CUB are fine-grained datasets, which may have a high demand for specialization ability. SUN consists of 14,340 images from 717 scene classes with 102 attributes. CUB contains 11,788 images from 200 bird species with 312 attributes. aPY and AwA2 are coarse-grained datasets, comprising 15,339 images from 32 classes with 64 attributes, and 37,322 images from 50 diverse animals with 85 attributes, respectively. We adopt Proposed Split (PS)~\cite{xian2019zero} to split datasets into seen and unseen classes. Our competitors also use the Proposed Split. 

Almost all compared methods use ResNet101 as the backbone net or use ResNet101-extracted 2048D features as visual embeddings. As such, we use ResNet101~\cite{he2016deep} as our Backbone Net, and last 9 layers of ResNet101 as tail layers for BSNet. $f_g$, $f_{b,i}$, and $f_s$ consist of one or two fully-connected layers. $w_i^d$ are randomly-initialized parameters that have gradients and can be optimized. We employ SGD~\cite{bottou2010large} as the optimizer, with a learning rate of $10^{-3}$. We adopt the 10-way, 5-shot, and 3-query setting to sample data for episodic meta-training of GNet. For BSNet, we set sub-module number $N=10$, scaling factor $s_{max}=5$, and max epoch $\mathit{eph}_{max}=100$. For GZSL, the Calibrated Stacking factor $\delta$ is set to 0.2, 0.6, 0.9, and 0.9 for SUN, CUB, aPY, and AwA2, respectively. $\eta$ is selected from $\{1e^{-4},1e^{-3},1e^{-2},1e^{-1},1\}$. 

We implement our algorithms under the Pytorch 1.7.0 framework. The compiler environment is GCC 7.3.0. All algorithms are running in the system of Linux 3.10.0, and based on the GPU of GP102 TITANX. The Cuda version is 10.0.130. Our model consists of two branches: GNet and BSNet. Get uses the ResNet 101 as backbone net, followed by $f_g$ to project visual embedding to attribute embeddings. $f_g$ is a one-layer Fully-Connected Network (FCN), with input size as 2048D and output size the same as attribute dimension. BSNet is composed of $N$ sub-modules and utilizes $f_s$ to combine the outputs of all sub-modules. Each sub-module has the same structure. The input image is first passed through the backbone net to get the 2048D visual embedding. Note that all sub-modules share most layers in the backbone net to save computation cost but adopt different tail layers for different output. Then the instance-level balance generator $f_{b,i }$ takes the generalized and specialized embeddings as input and outputs a 2048D embedding. $f_{b,i}$ is also one-layer FCN with input size as 4096D and output size as 2048D. Then the dataset-level balance is performed by $w^d_i$, a one -dimensional and differentiable weight initialized with random values. All balanced embeddings are concatenated to fed into $f_g$. $f_g$ has two fully-connected layers: $f_{g}=<FC(2048*N,2048)-FC(2048,d)-Dropout(0.4)>$, where $d$ is the dimension of attribute.

\subsection{Comparisons with SOTA}\label{exp main}

We compare BGSNet with 12 and 14 state-of-the-art methods for ZSL and GZSL settings, respectively. Our method is an end-to-end model focusing on improving image representation. It is fair to compare with other end-to-end methods. But to comprehensively review our performance gain, we further compare it with other generative methods, which synthesize unseen samples, and discriminative methods that find the best projection between visual and semantic space. For the compared methods, we use their reported results and pick up versions that do not use auxiliary side information, e.g., the without-group-information performance of APN~\cite{xu2020attribute}. 

\begin{figure*}[htb]
    \centering
    \subfloat[SUN]{\includegraphics[width=0.24\textwidth]{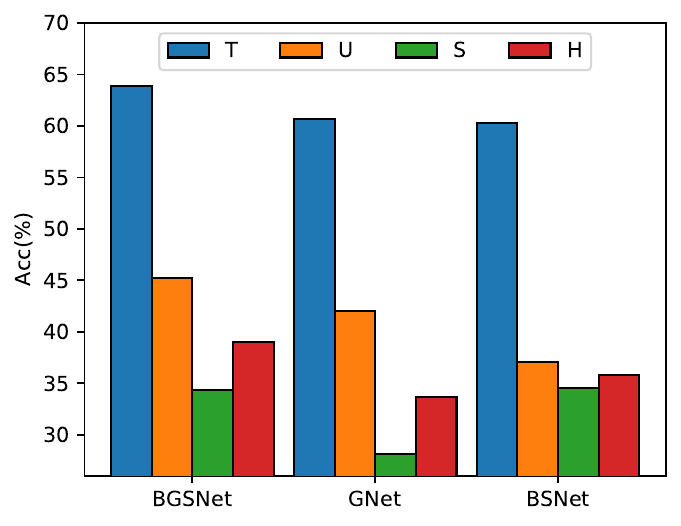}}\hfil 
    \subfloat[CUB]{\includegraphics[width=0.24\textwidth]{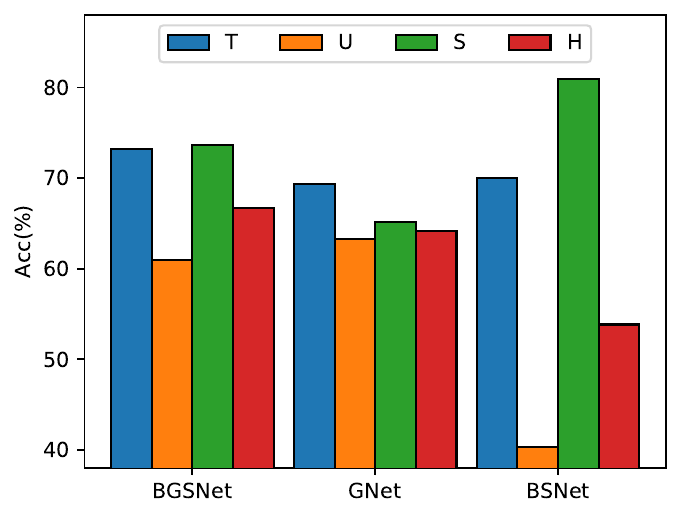}}\hfil 
      \subfloat[aPY]{\includegraphics[width=0.24\textwidth]{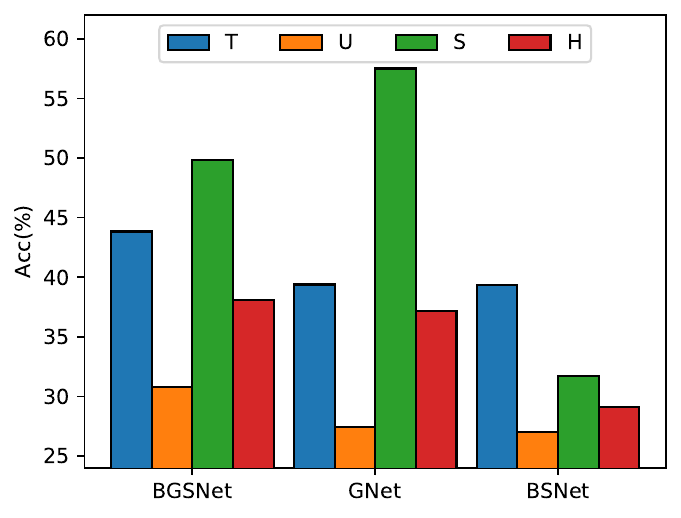}}\hfil 
      \subfloat[AwA2]{\includegraphics[width=0.24\textwidth]{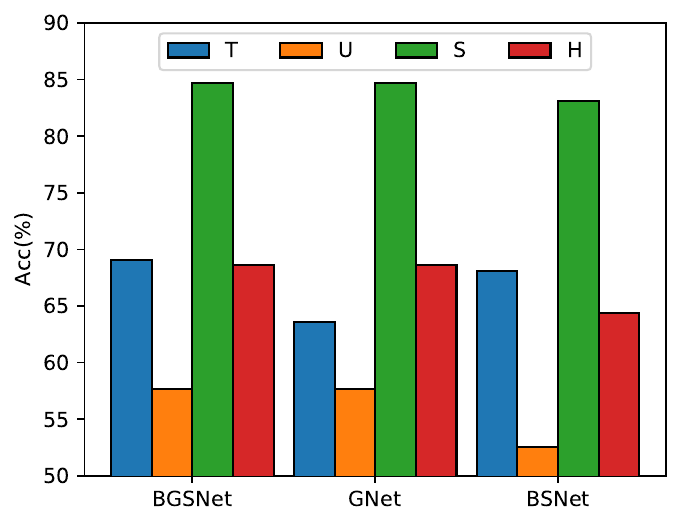}}\hfil 
    \\
      \subfloat[SUN]{\includegraphics[width=0.24\textwidth]{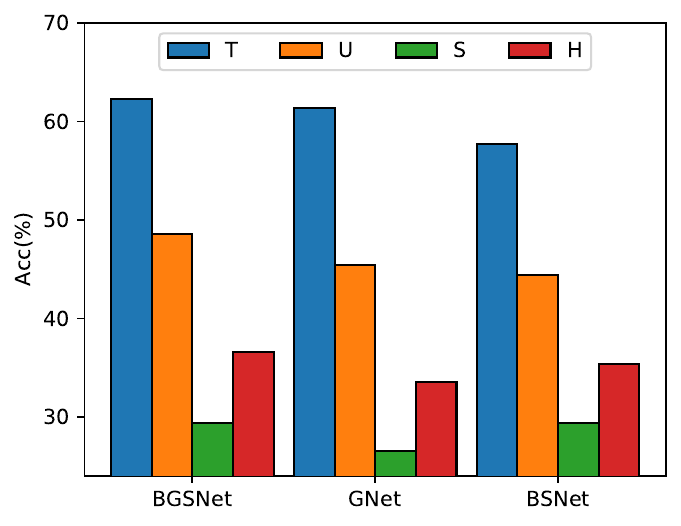}}\hfil 
    \subfloat[CUB]{\includegraphics[width=0.24\textwidth]{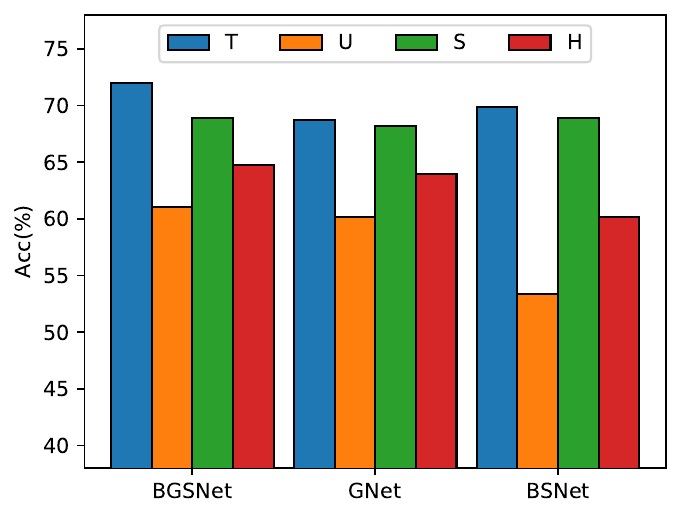}}\hfil 
    \subfloat[aPY]{\includegraphics[width=0.24\textwidth]{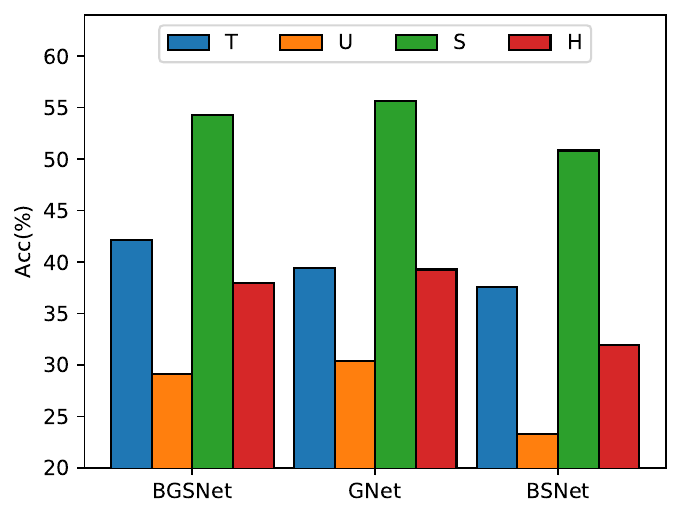}}\hfil 
    \subfloat[AwA2]{\includegraphics[width=0.24\textwidth]{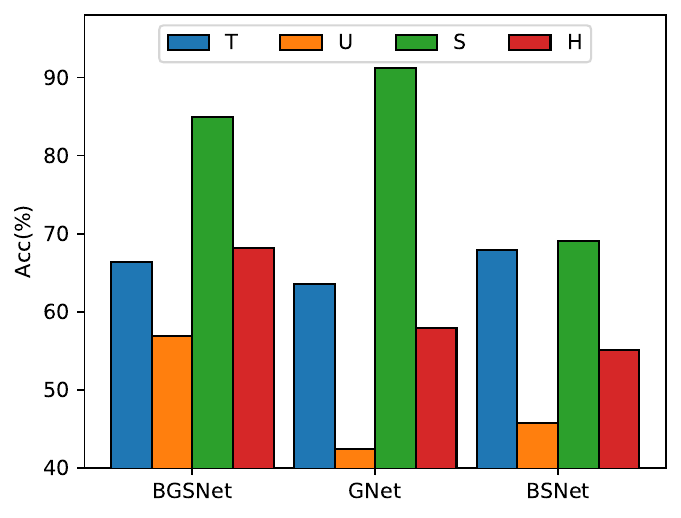}}\hfil 
    \\
    \centering
    \subfloat[SUN]{\includegraphics[width=0.24\textwidth]{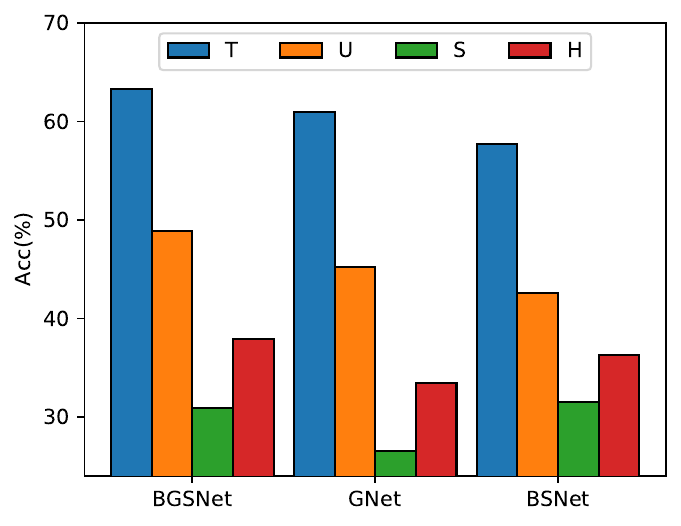}}\hfil 
    \subfloat[CUB]{\includegraphics[width=0.24\textwidth]{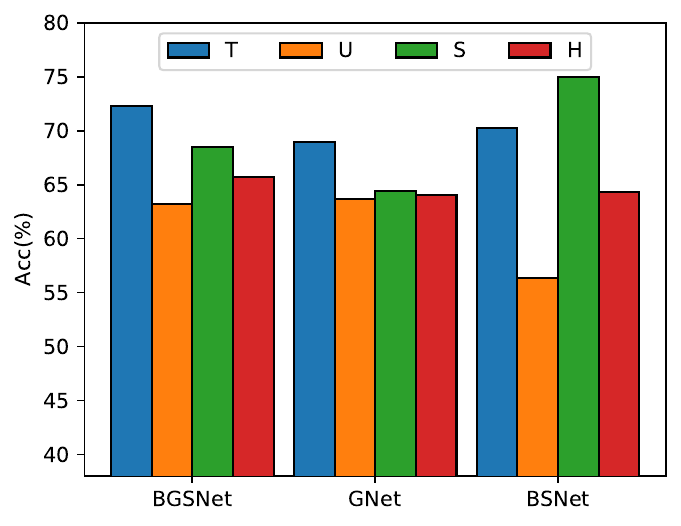}}\hfil 
    \subfloat[aPY]{\includegraphics[width=0.24\textwidth]{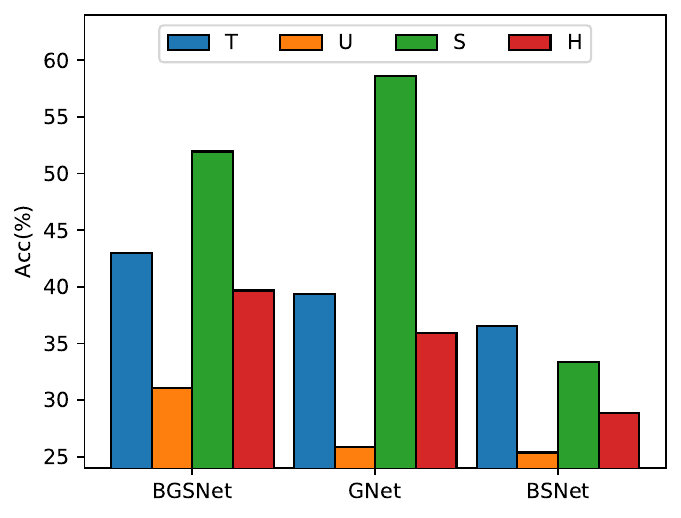}}\hfil 
    \subfloat[AwA2]{\includegraphics[width=0.24\textwidth]{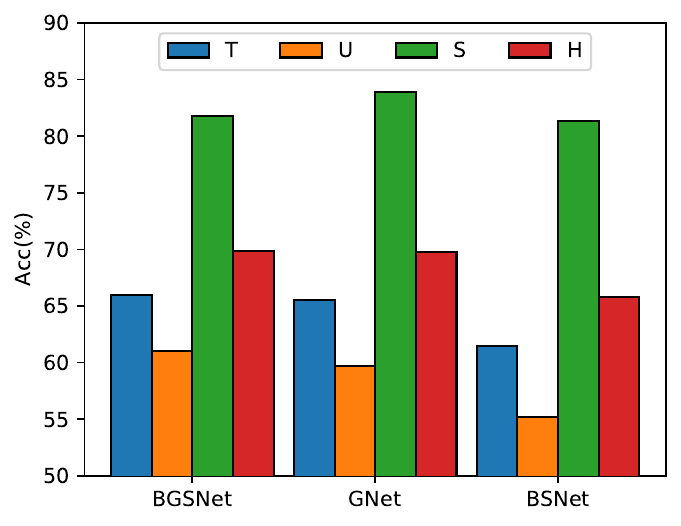}}\hfil 
  \hfil 
\caption{\textcolor{black}{Network module ablation of BGSNet on SUN, CUB, aPY, and AwA2, respectively. (a)-(d) Comparing BGSNet with two branches: GNet and BSNet. (e)-(h) Comparing BGSNet's variant w/o instance-level balance with GNet and BSNet. (i)-(l) Comparing BGSNet's variant that trained in parallel style with GNet and BSNet.}}
\label{network ablation}
\end{figure*}

For \textbf{ZSL}, we adopt the average per-class Top-1 (T) accuracy for evaluation to eliminate influence of class imbalance. Results are shown in Table~\ref{table zsl}. As can be seen, our BGSNet achieves comparable or better results than SOTA on all datasets and consistently outperforms other end-to-end methods by a large margin. In particular, BGSNet yields 2.4\%, 1.8\%, 4.6\%, and 0.3\% improvements over the second-best end-to-end methods. Comparing BGSNet with AMAZ~\cite{li2021attribute}, which also adopts meta-learning to improve the generalization ability, we can find that our model shows superior performance on fine-grained datasets, i.e., SUN and CUB. On the other side, our model performs much better than other attentive models, e.g., SGMA~\cite{zhu2019semantic}, AREN~\cite{xie2019attentive}, and APN~\cite{xu2020attribute}, on coarse-grained datasets, i.e., aPY and AwA2, whose classes are more diverse. In other words, our BGSNet can learn image representations that generalize better to unseen classes than attentive models with good specialization ability and contain more discriminative information than meta models which strengthens generalization ability. It indicates that our model can take advantage of both abilities, and the proposed balance can coordinate the two abilities based on instance and dataset characteristics. \textcolor{black}{We also report GNet without episodic meta-training in the table as Baseline. We can find that our model significantly outperforms the Baseline, validating our model's effectiveness.} 

For \textbf{GZSL}, we evaluate the methods with the average per-class Top-1 accuracy of seen classes ($U$), unseen classes ($S$), and their harmonic mean ($H=\frac{2US}{U+S}$). Among them, $H$ is the key criteria since it considers both seen and unseen classes. We summarize the results in Table~\ref{table gzsl}. From Table~\ref{table gzsl}, we can find that our model achieves impressive gains compared to its end-to-end counterparts for the harmonic mean: 1.4\%, 2.8\%, and 2.8\% improvements on SUN, aPY, and AwA2 respectively, and obtains second-best results on CUB. For GZSL, focusing on only one aspect of $U$ or $S$ may result in decreased $H$, e.g., LFGAA~\cite{Liu_2019_ICCV} achieves the best performance on AwA2 for $U$, but performs inferior to our BGSNet for $H$. This indicates that LFGAA improves generalization ability, leading to the degraded classification ability. On the contrary, our BGSNet can perform well on both $U$ and $H$. The results demonstrate that BGSNet can well combine and balance the generalization and specialization abilities when classifying images, e.g., reducing specialization ability by decreasing weights of BSNet when transferring knowledge to unseen classes, or vice versa.

\begin{figure*}[htb]
\centering
      \subfloat[]{\includegraphics[width=0.24\textwidth]{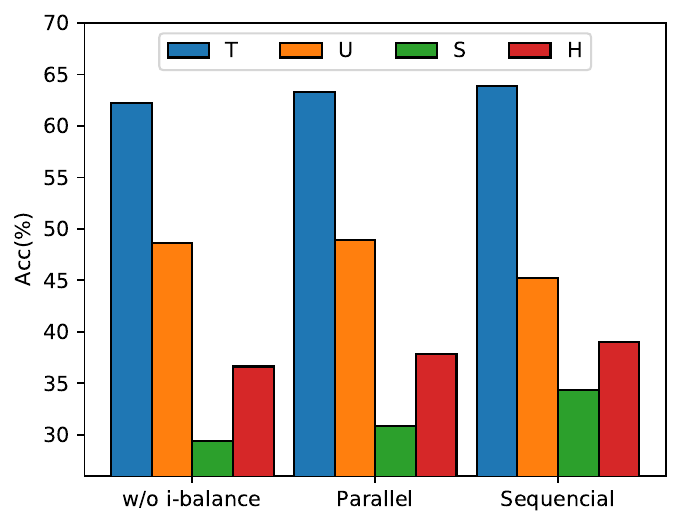}}\hfil 
      \subfloat[]{\includegraphics[width=0.24\textwidth]{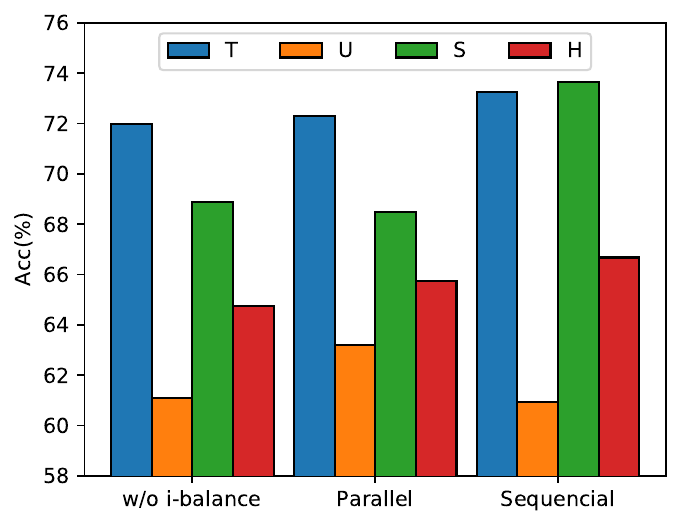}}\hfil 
      \subfloat[]{\includegraphics[width=0.24\textwidth]{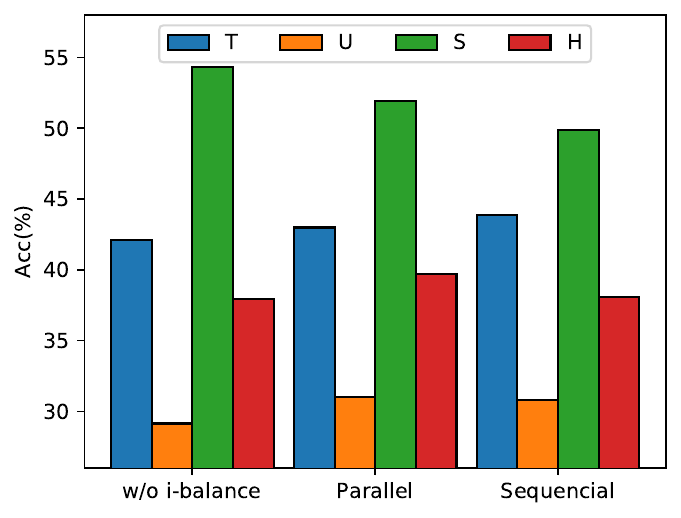}}\hfil 
      \subfloat[]{\includegraphics[width=0.24\textwidth]{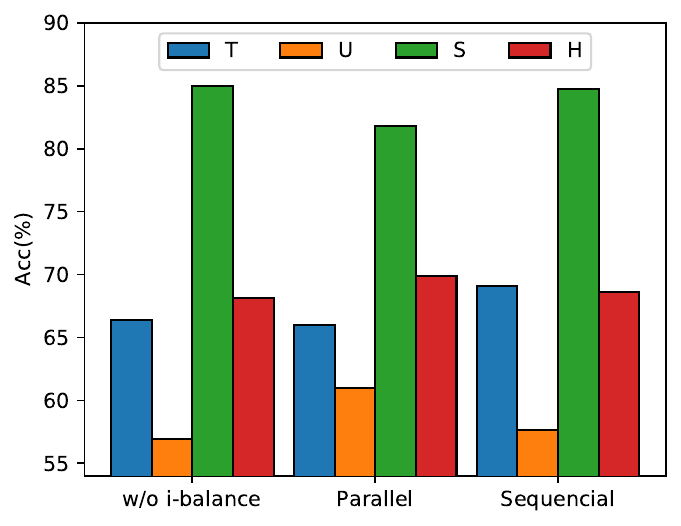}}\hfil 
  \hfil 
\caption{ Ablations of instance-level balance  (i-balance for short in the figures) and training styles on (a) SUN, (b) CUB, (c) aPY, and (d) AwA2, respectively. }
\label{training ablation}
\end{figure*}

\begin{figure*}[htb]
\centering
    \subfloat[]{\includegraphics[width=0.25\textwidth]{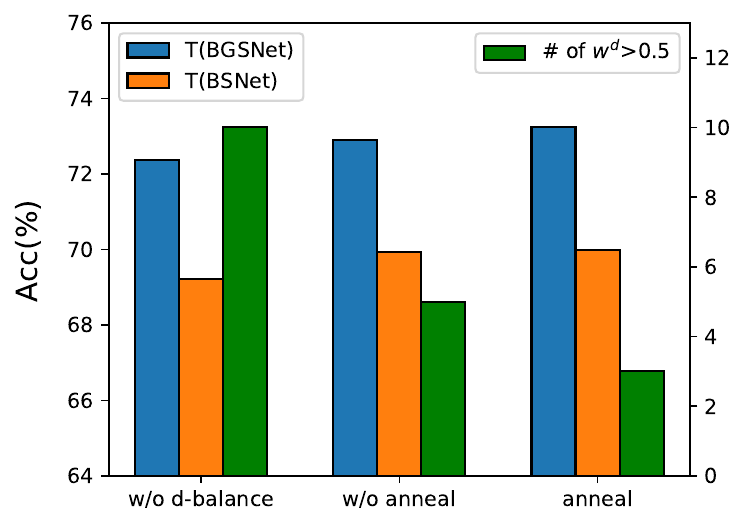}}\hfil 
    \subfloat[]{\includegraphics[width=0.22\textwidth]{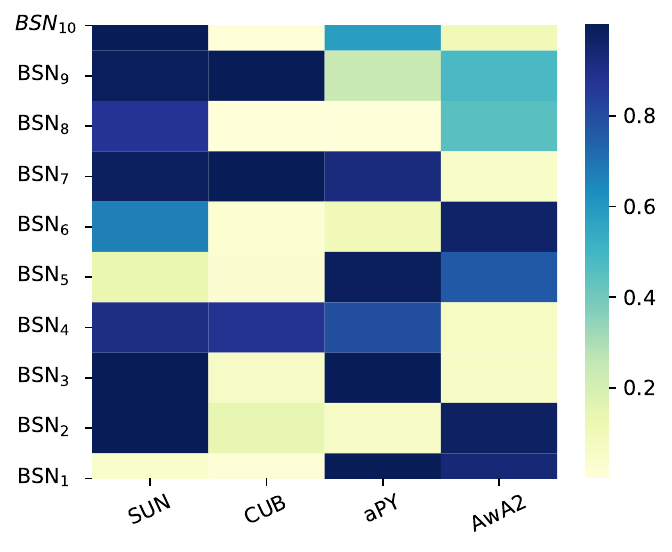}}\hfil 
    \subfloat[]{\includegraphics[width=0.22\textwidth]{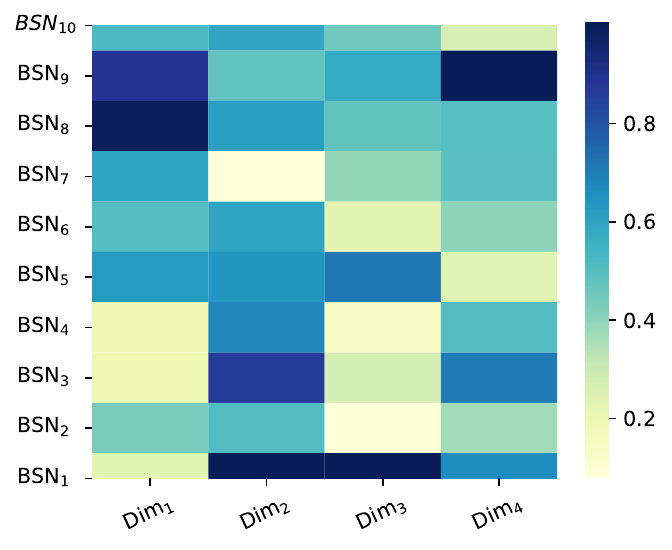}}\hfil 
    \subfloat[]{\includegraphics[width=0.22\textwidth]{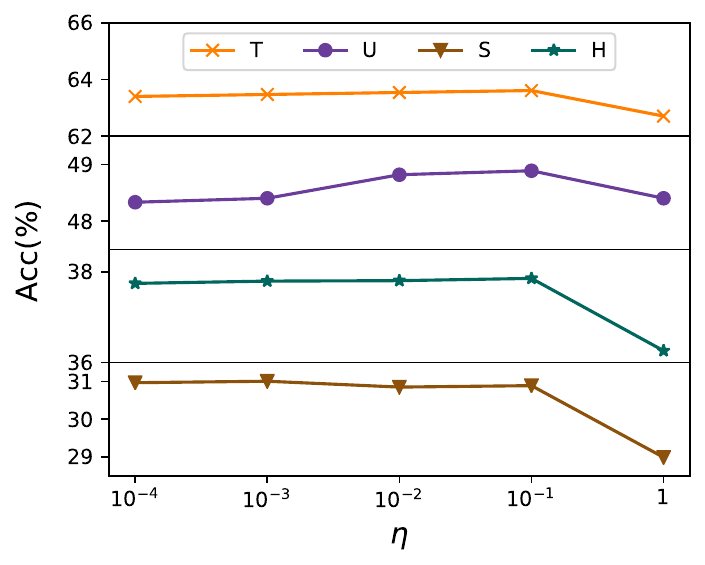}}\hfil 
\caption{Balance analysis. (a)-(b) Ablation of dataset-level balance. (a) CUB dataset balance. (b) Distributions of $w^d$ on SUN, CUB, aPY, and AwA2. (c)-(d) Analysis of instance-level balance. (c) $w^t$ on SUN. (d) $\eta$ on SUN.}
\label{di_analysis}
\end{figure*}

\subsection{Ablation Study}

\subsubsection{Necessity of Two Branches} We compare BGSNet (trained in \textit{Sequential}) with its two branches: GNet and BSNet, each focusing only on generalization or specialization abilities. Figures~\ref{network ablation} (a)-(d) show the results. We can observe that BGSNet outperforms GNet and BSNet on all datasets for both ZSL and GZSL settings, indicating that combining GNet and BSNet is necessary and can improve the model's adaptability to unseen classes and differentiation between seen classes at the same time. 
We also find that GNet always outperforms BSNet in $U$, consistent with our expectation that generalization is more critical when generalizing to unseen classes. However, BSNet is not always superior to GNet in terms of $S$. One possible reason is that we use $L_{div}$ to restrict BSNet only to learn novel knowledge that GNet ignores so that BSNet may lack some basic knowledge for classification. 

We also compare BGSNet w/o instance-level balance (trained in an end-to-end manner) and BGSNet trained in \textit{Paralled}  with their two branches: GNet and BSNet. Figure~\ref{network ablation} (e)-(l) show the results. We can observe that the two variants of BGSNet exhibit similar patterns as BGSNet. They all outperform GNet and BSNet on most datasets for both ZSL and GZSL settings, indicating the necessity of combining GNet and BSNet. 

\subsubsection{Benefits of Instance-level Balance}
We then evaluate the effects of the instance-level balance. We compare BGSNet with its variant without instance-level balance and the variants that adopt different training strategies. Figures~\ref{training ablation} (a)-(d) report the results. We find that variant without instance-level balance obtains the worst performance on most datasets for both settings. The reason lies in that, without instance-level balance, BGSNet relies only on dataset-level balance to weigh GNet and BSNet, which is not as sufficient and suitable for a specific image as instance-level balance. Besides, our proposed $L_{div}$ can optimize BSNet to capture novel knowledge neglected by GNet, and improve specialization diversity. As such, the absence of $L_{div}$ also impairs the performance. 

When comparing two training strategies, we observe that \textit{Sequential} perform better than \textit{Parallel} on SUN and CUB, while worse on aPY and AwA2 for GZSL. This may be caused by the dataset characteristics. As discussed in Section~\ref{sec:training}, training in parallel results in better-cooperated instance- and dataset- level balance and unstable optimization of BSNet. The unstable optimization may impair its specialization ability, proved by \textit{Sequential}'s superiority over \textit{Parallel} in $S$ on most datasets. 
aPY and AwA2 are datasets with highly diverse classes and thus may have higher balance demands, while SUN and CUB need specialization ability more as they contain highly similar classes, leading to  different preference of \textit{Sequential} and \textit{Parallel}. 

\begin{figure*}[htb]
    \centering
    \subfloat[]{\includegraphics[width=0.24\textwidth]{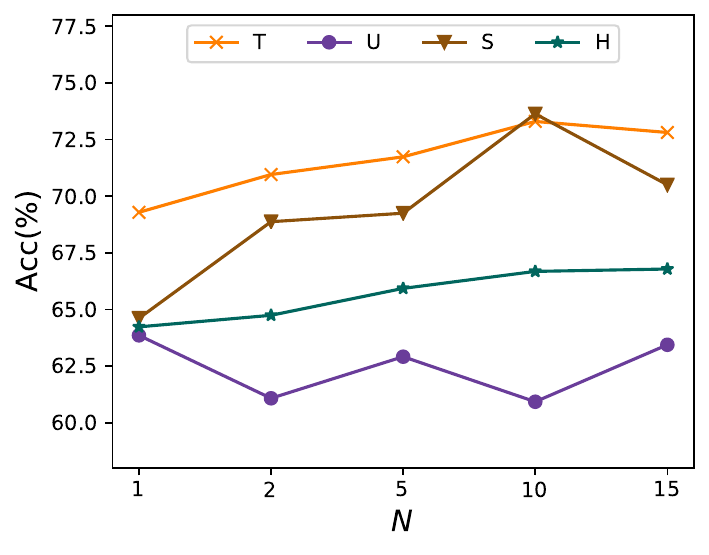}}\hfil 
    \subfloat[]{\includegraphics[width=0.24\textwidth]{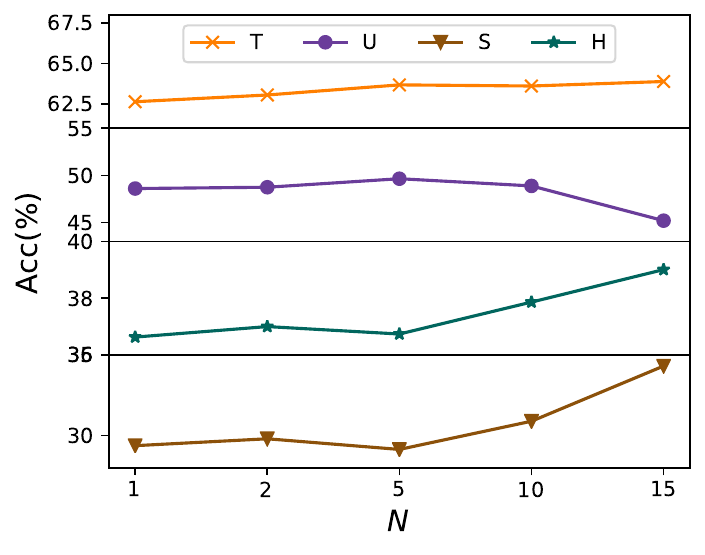}}\hfil 
    \subfloat[]{\includegraphics[width=0.24\textwidth]{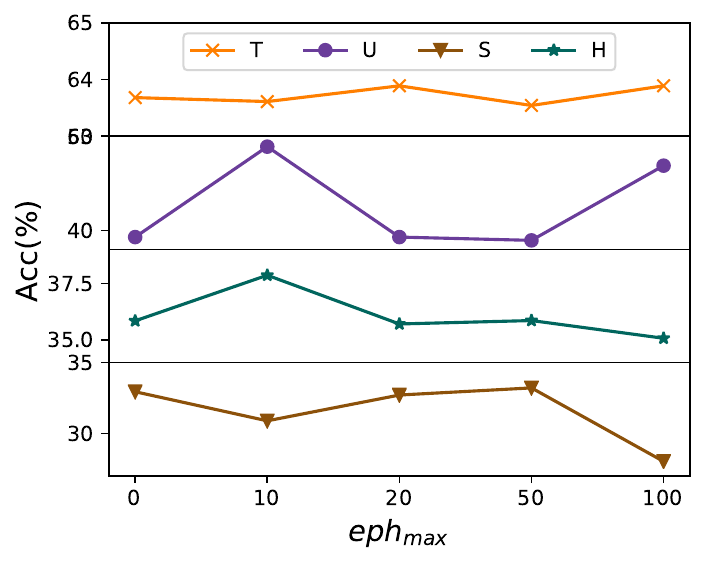}}\hfil 
    \subfloat[]{\includegraphics[width=0.24\textwidth]{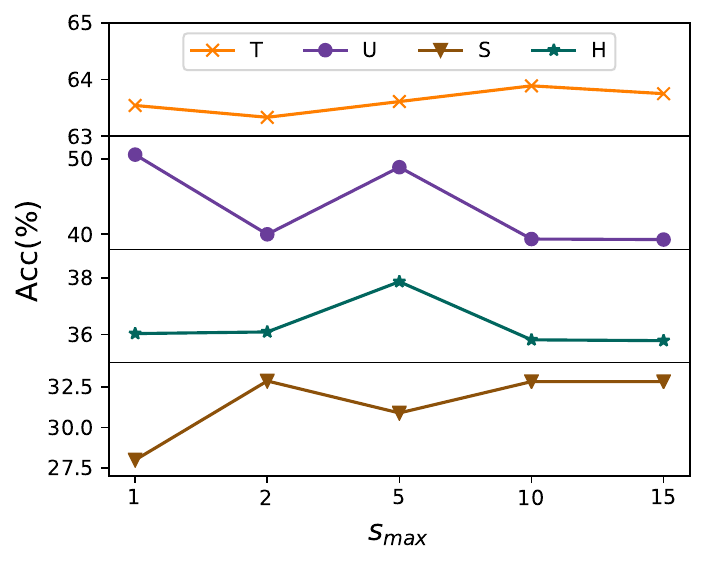}}\hfil 
\caption{Hyper parameter analysis of BGSNet. (a) $N$ analysis on CUB. (b) $N$ analysis on SUN. (c) $\mathit{eph}_{max}$ analysis on SUN. (d) $s_{max}$ analysis on SUN.}
\label{snet_analysis}
\end{figure*}

\begin{figure*}[htbp]
    \centering
    \subfloat[]{\includegraphics[width=0.23\textwidth]{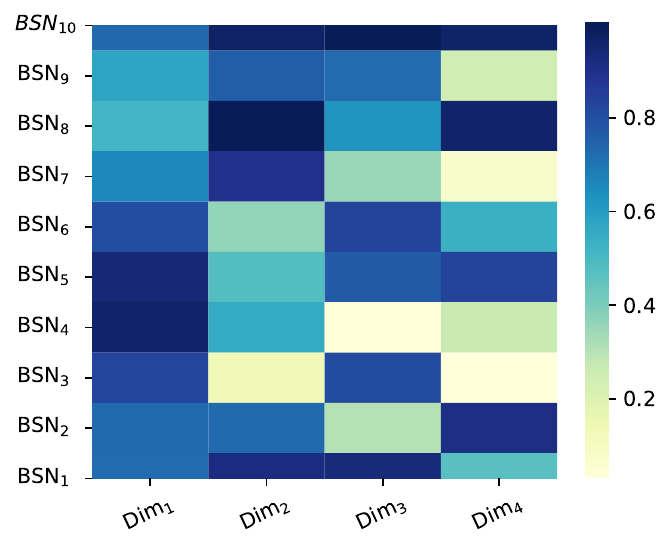}}\hfil 
    \subfloat[]{\includegraphics[width=0.23\textwidth]{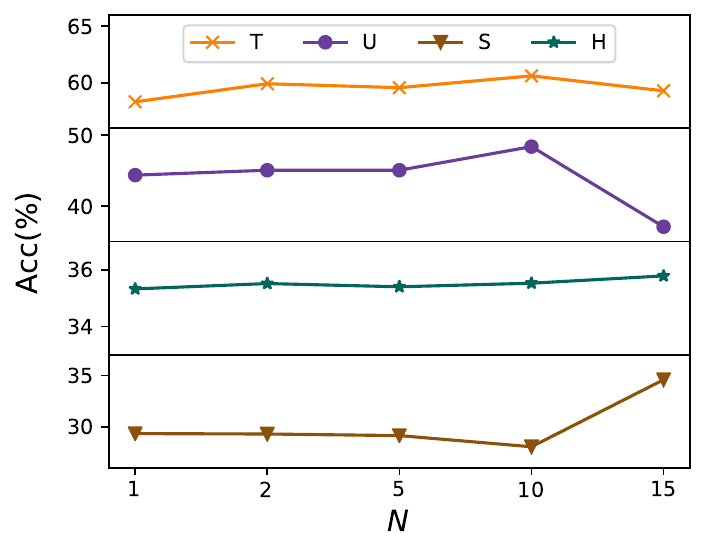}}\hfil 
    \subfloat[]{\includegraphics[width=0.23\textwidth]{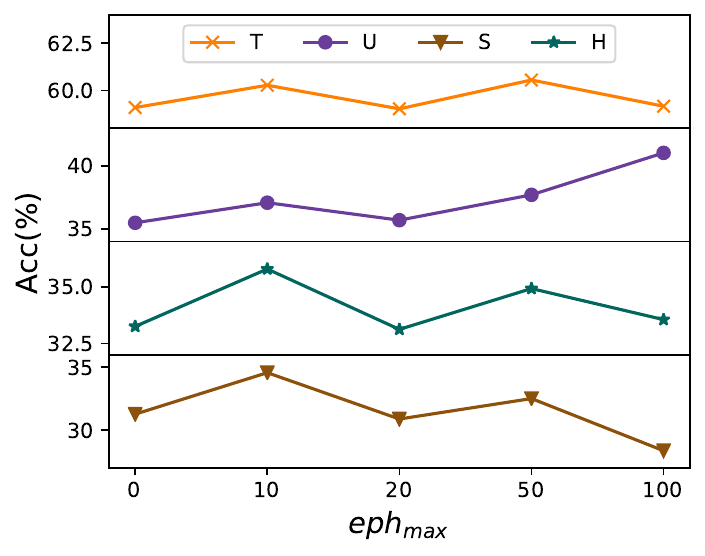}}\hfil 
    \subfloat[]{\includegraphics[width=0.23\textwidth]{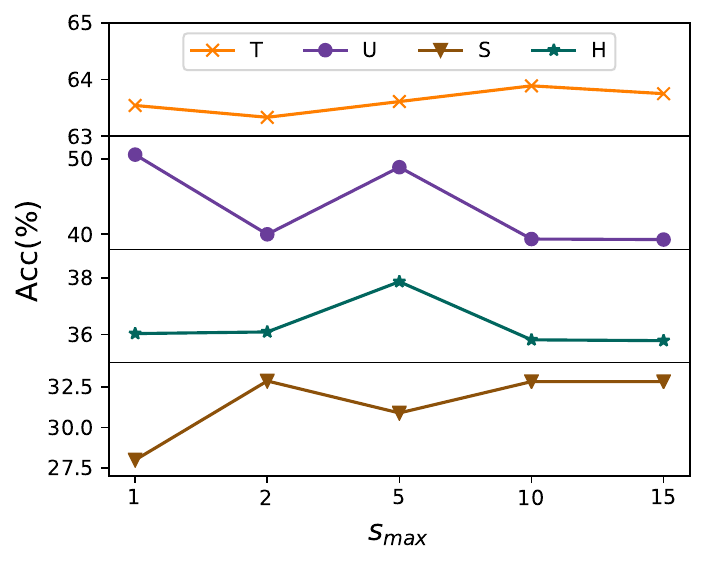}}\hfil 
\caption{Other Analysis. (a) Distribution analysis of $w^t$ on CUB. (b)-(d) Hyper parameter analysis of BSNet on SUN. (b) $N$ on SUN. (c) $\mathit{eph}_{max}$ on SUN. (d) $s_{max}$ on SUN.}
\label{supp_snet_analysis}
\end{figure*}

\subsubsection{Benifits of Dataset-level Balance}
We compare BGSNet with variant without dataset-level balance and variant without linear annealing. As shown in Figure~\ref{di_analysis} (a), including dataset-level balance improves the performance of both BGSNet and BSNet. Dataset-level balance can enhance important sub-modules by generating dataset-specific weights, and thus benefits BSNet, and better weighs GNet and BSNet. We also report the number of sub-modules with $w^d_i>0.5$, and find that the annealing schedule can effectively reduce required sub-modules. BSNet and BGSNet even perform better with fewer sub-modules , demonstrating that dataset-level balancing in annealing schedule can find the optimal structure of BSNet.

\subsection{Balance analysis}

\subsubsection{Dataset-level Balance} We draw the distribution of $\{w^d_i\}^N_1$ in Figure~\ref{di_analysis} (b). SUN needs most sub-modules in BSNet, indicating its high demand of specialization ability. However, another fine-grained dataset, CUB, needs fewer sub-modules than AwA2 and aPY. The reason may be that images in CUB are all about birds, and to discriminate birds, we only need to focus on a few areas such as wings, heads, bellies, etc.

\subsubsection{Instance-level Balance on SUN}  We randomly sample 4 dimensions from $K$ channels of $\{w^t_i\}$ and draw their distribution in Figure~\ref{di_analysis} (c). The weights are diverse across sub-modules and channels, proving that our proposed diversity loss $L_{div}$ can optimize the instance-level balance to meet our two requirements, and thereby, the two functions of $w^t_i$ are fulfilled. We further vary ratio $\eta$ of $L_{div}$ and show results on SUN in Figure~\ref{di_analysis} (d). $L_{div}$ is much larger than $L_{s}$. Thus the performance is stable when $\eta$ is small and decreases when $\eta=1$. Analysis for CUB is provided in~\ref{cub analysis}.

\begin{table}[htb]
\centering
\caption{Hyper parameter analysis of GNet. The performance are evaluated by the average per-class Top-1 accuracy (\%). Best results are in bold.}
\begin{tabular}{lllll}
\toprule
\#-way~~& \#-shot~~ & \#-query~~ & AwA2~~ & CUB~~ \\
\midrule
-&-&-& \textcolor{black}{65.5} & 68.7 \\
5 & 3 & 3 & 64.0 & 69.1 \\ 
5 & 5 & 3 & 65.5 & 69.3 \\
5 & 5 & 5 & 63.7 & 69.0 \\
10 & 3 & 3 & 65.8 & 68.9 \\
10 & 5 & 3 & \textbf{67.8} & \textbf{70.3} \\
10 & 5 & 5 & 64.7 & 69.0 \\
15 & 3 & 3 & 65.6 & 68.6 \\
15 & 5 & 3 & 65.9 & 69.1 \\
15 & 5 & 5 & 65.7 & 69.5 \\ 
\bottomrule
\end{tabular}
\label{tab:meta_abaltion}
\end{table}

\subsection{Hyper parameters and more experiments}

\subsubsection{Episodic Setting of GNet} We try different settings for episodic meta-learning and compare with using the random sampling training strategy. Table~\ref{tab:meta_abaltion} shows their T accuracy for ZSL. Episodic meta training improves the model performance than random sampling. The model gets the highest accuracy when sampling tasks in 10-way, 5-shot, and 3-query.

\subsubsection{$N$, $eph_{max}$ and $s_{max}$ in BGSNet} \label{exp hyper}We analyze BGSNet's  sensitivity to hyper-parameters, i.e., $N$, $eph_{max}$ and $s_{max}$. Figure~\ref{snet_analysis} shows the results. The key criterion T and H increase as $N$ increases, and achieves best results at 10 for CUB and 15 for SUN. The maximum epoch $eph_{max}$ needs to be less than 50 for SUN. The scaling factor $s_{max}$ is best at 5.

\subsubsection{Distribution analysis of $w^t$ on CUB}\label{cub analysis} For CUB, we randomly sample 4 dimensions from $K$ channels of $\{w^t_i\}$ and draw their distribution in Figure~\ref{supp_snet_analysis} (a). The weights are diverse across sub-modules and channels, proving the effectiveness of our proposed diversity loss $L_{div}$. 

\subsubsection{Hyper parameter analysis of BSNet on SUN}
We also analyze BSNet's  sensitivity to hyper-parameters. Figures~\ref{supp_snet_analysis} (b)-(d) show the results. Unlike the results of BGSNet in~\ref{exp hyper}, this hyper-analysis only studies the performance of BSNet. Thus it excludes the influence of GNet, and cannot observe how well GNet and BSNet cooperate as hyper-parameters vary. The key criteria T and H are stable as $N$ increases. The maximum epoch $eph_max$ achieves the best results at 10. The scaling factor $s_max$ is best at 5. 

\subsubsection{\textcolor{black}{Hyper parameter analysis of $\delta$ on AwA2}}\textcolor{black}{ We also analyze how $\delta$ affects the GZSL performance. Table~\ref{delta} show the results.  We can find that as $\delta$ increases, $U$ increases, $S$ decreases, and $H$ increases. When $\delta$ is set to 1.0, all images are predicted as unseen classes,  $U$ gets the highest accuracy, and $S$ is 0.}

\begin{table}[]
\centering
\color{blue}{\caption{Hyper parameter analysis of $\Delta$}}
\label{delta}
\begin{tabular}{l|cccccc}
\toprule
$\delta$ & 0.1  & 0.3  & 0.5  & 0.7  & 0.9 & 1.0 \\
\midrule
 U        & 9.2  & 20.3 & 39.1 & 51.4 & 61.0 & 66.8 \\
 S        & 95.1 & 94.4 & 92.0 & 88.5 & 81.8 & 0.0 \\
 H        & 16.8 & 33.4 & 54.9 & 65.0 & 69.9 & 0.0\\
\bottomrule
\end{tabular}
\end{table}

\section{Related Work}
\subsection{Zero-Shot Learning (ZSL)} A typical solution~\cite{li2022entropy,li2021attribute,ye2021alleviating,ye2021disentangling,ye2022learning,guo2017zero} to ZSL, called \textbf{discriminative methods}, is to project visual and semantic embeddings to a common space, and then perform nearest neighbor search in the space for classification. The common space can be the semantic~\cite{ye2019progressive,imrattanatrai2019identifying,guo2020novel}, visual~\cite{zhang2017learning,shigeto2015ridge,shen2021spherical}, or a new latent space~\cite{morgado2017semantically,jiang2019transferable}. For example, Li et al.~\cite{li2020learning} leverage the mutual information and entropy to learn modality-invariant embedding. TCN~\cite{jiang2019transferable} designs a contrastive network to exploit the class similarities for robust embeddings.

Recently, some studies focus on \textbf{generative methods} to synthesize samples for unseen classes, converting ZSL to a traditional supervised learning problem~\cite{chi2019zero,zhang2019triple,gao2020zero,li2021attribute,liu2021task}. DADN~\cite{chi2019zero} is a dual adversarial distribution network with two generative branches. The two branches promote each other, and learn the inter-media correlation and semantic ranking respectively.
Liu et el.~\cite{liu2021task} propose to first align the distributions of visual embeddings belonging to the same task, and then generate samples based on the task-aligned embedding. 

\textcolor{black}{Besides, some modern works further enhance ZSL by adopting \textbf{ensemble-based networks}~\cite{changpinyo2017predicting,ding2017low,ye2019progressive,chen2020rethinking}. For example, Ye~\cite{ye2019progressive} proposes to ensemble multiple image classification functions, and progressively train and augment the model with the most confident pseudo-labels predicted by the ensemble network. Chen~\cite{chen2020rethinking} generate multiple visual patches from text descriptions, and uses them to ensemble diverse classifiers based on attentive weights and voting strategies.}

More related to our models are \textbf{end-end-to models}, which adopt backbone net, e.g., ResNet101, to learn image representation, and then based on the learned visual embeddings to perform classification by the discriminative strategy~\cite{Liu_2019_ICCV,xie2019attentive,xie2020region}, or generative strategy~\cite{xian2019f,chou2020adaptive,narayan2020latent,zhang2021modality}. LFGAA~\cite{Liu_2019_ICCV} integrate the backbone net with latent attribute attention to reduce semantic ambiguity.
Xian et al.~\cite{xian2019f} fine-tune the backbone net to get visual embeddings, combine VAE and GAN to form a generator, and generate samples for unseen classes. Our BGSNet adopts the discriminative strategy after learning image representations.

\subsection{Generalization and Specialization in ZSL} Reviewing the related works, we find that generalization and specialization are two essential abilities for ZSL. 
This is because, ZSL aims to classify unseen images based on knowledge learned from seen classes, and can be disentangled into two tasks, named knowledge transfer and image classification. 
The two tasks require the two abilities respectively. Many efforts are made to improve the generalization~\cite{li2021attribute,liu2021task,liu2021isometric} or specialization abilities~\cite{xie2019attentive,zhu2019semantic,xu2020attribute,liu2021rethink}. Liu et al.~\cite{liu2021isometric} adapt episodic meta-learning to propagate the attribute and visual prototypes to improve generalization and relieve class imbalance. RSR~\cite{liu2021rethink} proposes a spiral revising framework to enhance semantic specialization under the guidance of a reinforced attribute-grouping module. Xie et al.~\cite{xie2019attentive} utilize multiple attention maps to increase visual specialization. Our model, like ~\cite{liu2021isometric}, uses meta-learning for better generalization and proposes the instance-level balance and diversity loss for diverse specialization.

\subsection{Summary} Our model differs from previous studies in three aspects. 1) Our model equips both generalization and specialization abilities and balances the two abilities according to instance- and dataset-characteristics. However, most existing methods only focus on one of them, let alone balance them. 2) We design a novel diversity loss, which is dynamic with a self-calculated margin to adjust the optimization purpose during training. 3)We propose to update the dataset-level weight in a linear annealing schedule to find optimal network structure for a specific dataset.

\section{Conclusion}
This paper proposes a novel BGSNet for zero-shot learning and generalized zero-shot learning. 
We develop two branches, GNet and BSNet, to learn transferable knowledge and extract discriminative features, respectively. The balance between the two branches is accomplished at the instance- and dataset-level. 
Instance-level balance is optimized by a newly proposed diversity loss to reduce redundancy and boost diversity across sub-modules of BSNet. In addition, the dataset-level balance can also play the role of network architecture search to find optimal dataset-specific structures for BSNet. 
Extensive experiments show the effectiveness of our proposed methods in differentiating images and balancing the specialization and generalization abilities based on the instance and dataset types. In the future, we plan to extend our method in more real-world applications that may suffer from the imbalanced network learning abilities, e.g., human activity recognition and natural language processing.

\bibliographystyle{IEEEtran}
\bibliography{bio}

 





\end{document}